# Convergence Rates of Biased Stochastic Optimization for Learning Sparse Ising Models

**Jean Honorio**                                                                                                                  JHONORIO@CS.SUNYSB.EDU
Stony Brook University, Stony Brook, NY 11794, USA

## Abstract

We study the convergence rate of *stochastic* optimization of *exact* (NP-hard) objectives, for which only biased estimates of the gradient are available. We motivate this problem in the context of learning the structure and parameters of Ising models. We first provide a convergence-rate analysis of *deterministic* errors for *forward-backward splitting* (FBS). We then extend our analysis to *biased stochastic* errors, by first characterizing a family of samplers and providing a high probability bound that allows understanding not only FBS, but also *proximal gradient* (PG) methods. We derive some interesting conclusions: FBS requires only a logarithmically increasing number of random samples in order to converge (although at a very low rate); the required number of random samples is the same for the deterministic and the biased stochastic setting for FBS and basic PG; accelerated PG is not guaranteed to converge in the biased stochastic setting.

## 1. Introduction

Structure learning aims to discover the topology of a probabilistic network of variables such that this network represents accurately a given dataset while maintaining low complexity. Accuracy of representation is measured by the likelihood that the model explains the observed data, while complexity of a graphical model is measured by its number of parameters.

One challenge of structure learning is that the number of possible structures is super-exponential in the number of variables. For Ising models, the number of parameters, the number of edges in the structure and the number of non-zero elements in the *ferro-magnetic coupling* matrix are equivalent measures of model complexity. Therefore a computationally tractable approach is to use sparseness promoting regularizers (Wainwright et al., 2006; Banerjee et al., 2008; Höfling & Tibshirani, 2009).

One additional challenge for Ising models (and Markov random fields in general) is that computing the likelihood of a candidate structure is NP-hard. For this reason, several researchers propose exact optimization of approximate objectives, such as $\ell_1$-regularized logistic regression (Wainwright et al., 2006), greedy optimization of the conditional log-likelihoods (Jalali et al., 2011), pseudo-likelihood (Besag, 1975) and a sequence of first-order approximations of the exact log-likelihood (Höfling & Tibshirani, 2009). Several convex upper bounds and approximations to the log-partition function have been proposed for maximum likelihood estimation, such as the log-determinant relaxation (Banerjee et al., 2008), the cardinality bound (El Ghaoui & Gueye, 2008), the Bethe entropy (Lee et al., 2006; Parise & Welling, 2006), tree-reweighted approximations and general weighted free-energy (Yang & Ravikumar, 2011).

In this paper, we focus on the stochastic optimization of the exact log-likelihood as our motivating problem. The use of *stochastic maximum likelihood* dates back to (Geyer, 1991; Younes, 1988), in which Markov chain Monte Carlo (MCMC) was used for approximating the gradient. For restricted Boltzmann machines (a very related graphical model) researchers have proposed a variety of approximation methods, such as variational approximations (Murray & Ghahramani, 2004), contrastive divergence (Hinton, 2002), persistent contrastive divergence (Tieleman, 2008), tempered MCMC (Salakhutdinov, 2009; Desjardins et al., 2010), adaptive MCMC (Salakhutdinov, 2010) and particle filtering (Asuncion et al., 2010).

Empirical results in (Marlin et al., 2010) suggests that stochastic maximum likelihood is superior to con-trastive divergence, pseudo-likelihood, ratio matching and generalized score matching for learning restricted Boltzmann machines, in the sense that it produces a higher test set log-likelihood, and more consistent classification results across datasets.

Learning sparse Ising models leads to the use of stochastic optimization with biased estimates of the gradient. Most work in stochastic optimization assumes the availability of unbiased estimates (Duchi & Singer, 2009; Duchi et al., 2010; Hu et al., 2009; Nemirovski et al., 2009). Additionally, other researchers have analyzed convergence rates in the presence of *deterministic* errors that do not decrease over time (d'Aspremont, 2008; Baes, 2009; Devolder et al., 2011) and show convergence up to a constant level. Similarly, Devolder (2012) analyzed the case of *stochastic* errors with *fixed* bias and variance and show convergence up to a constant level.

Notable exceptions are the recent works of Schmidt et al. (2011); Friedlander & Schmidt (2011); Duchi et al. (2011). Schmidt et al. (2011) analyzed *proximal-gradient* (PG) methods for *deterministic* errors of the gradient that decrease over time, for inexact projection steps and Lipschitz as well as strongly convex functions. In our work, we restrict our analysis to exact projection steps and do not assume strong convexity. Both assumptions are natural for learning sparse models under the $\ell_1$ regularization. Friedlander & Schmidt (2011) provides convergence rates in expected value for PG with *stochastic* errors that decrease over time in expected value. Friedlander & Schmidt (2011) proposes a growing sample-size strategy for approximating the gradient, i.e. by picking an increasing number of training samples in order to better approximate the gradient. In contrast, our work is for NP-hard gradients and we provide bounds with high probability, by taking into account the bias and the variance of the errors. Duchi et al. (2011) analyzed *mirror descent* (a generalization that includes forward-backward splitting) and show convergence rates in expected value and with high probability with respect to the mixing time of the sampling distribution. We argue that practitioners usually terminate Markov chains before properly mixing, and therefore we motivate our analysis for a controlled increasing number of random samples.

Regarding our contribution in optimization, we provide a convergence-rate analysis of *deterministic* errors for three different flavors of *forward-backward splitting* (FBS): robust (Nemirovski et al., 2009), basic and random (Duchi & Singer, 2009). We extend our analysis to *biased stochastic* errors, by first characterizing a family of samplers (including importance sampling and

MCMC) and providing a high probability bound that is useful for understanding the convergence of not only FBS, but also PG (Schmidt et al., 2011). Our analysis shows the bias/variance term and allow to derive some interesting conclusions. First, FBS for deterministic or biased stochastic errors requires only a logarithmically increasing number of random samples in order to converge (although at a very low rate). More interestingly, we found that the required number of random samples is the same for the deterministic and the biased stochastic setting for FBS and basic PG. We also found that accelerated PG is not guaranteed to converge in the biased stochastic setting.

Regarding our contribution in structure learning, we show that the optimal solution of maximum likelihood estimation is bounded (to the best of our knowledge, this has not been shown before). Our analysis shows provable convergence guarantees for finite iterations and finite number of random samples. Note that while consistency in structure recovery has been established (e.g. Wainwright et al. (2006)), convergence rates of parameter learning for fixed structures is up to now unknown. Our analysis can be easily extended to Markov random fields with higher order cliques as well as parameter learning for fixed structures by using a $\ell_2^2$ regularizer instead.

## 2. Our Motivating Problem

In this section, we introduce the problem of learning sparse Ising models and discuss its properties. Our discussion will motivate a set of bounds and assumptions for a more general convergence rate analysis.

### 2.1. Problem Setup

An *Ising model* is a Markov random field on binary variables with pairwise interactions. It first arose in statistical physics as a model for the energy of a physical system of interacting atoms (Koller & Friedman, 2009). Formally, the probability mass function (PMF) of an Ising model parameterized by $\boldsymbol{\theta} = (\mathbf{W}, \mathbf{b})$ is defined as:

$$p_{\boldsymbol{\theta}}(\mathbf{x}) = \frac{1}{\mathcal{Z}(\mathbf{W}, \mathbf{b})} e^{\mathbf{x}^\mathsf{T} \mathbf{W} \mathbf{x} + \mathbf{b}^\mathsf{T} \mathbf{x}} \qquad (1)$$

where the domain for the binary variables is $\mathbf{x} \in \{-1, +1\}^N$, $\mathbf{W} \in \mathbb{R}^{N \times N}$ is symmetric with zero diagonal, $\mathbf{b} \in \mathbb{R}^N$ and partition function is defined as $\mathcal{Z}(\mathbf{W}, \mathbf{b}) = \sum_{\mathbf{x}} e^{\mathbf{x}^\mathsf{T} \mathbf{W} \mathbf{x} + \mathbf{b}^\mathsf{T} \mathbf{x}}$. For clarity of the convergence rate analysis, we also define $\boldsymbol{\theta} \in \mathbb{R}^M$ where $M = N^2$.

In the physics literature, $\mathbf{W}$ and $\mathbf{b}$ are called *ferromagnetic coupling* and *external magnetic field* respec-tively. $\mathbf{W}$ defines the topology of the Markov random field, i.e. the graph $\mathcal{G} = (\mathcal{V}, \mathcal{E})$ is defined as $\mathcal{V} = \{1, \ldots, N\}$ and $\mathcal{E} = \{(n_1, n_2) \mid n_1 < n_2 \wedge w_{n_1 n_2} \neq 0\}$. It is well known that, for an Ising model with arbitrary topology, computing the partition function $\mathcal{Z}$ is NP-hard (Barahona, 1982). It is also NP-hard to approximate $\mathcal{Z}$ with high probability and arbitrary precision (Chandrasekaran et al., 2008).

The number of edges $|\mathcal{E}|$ or equivalently the cardinality (number of non-zero entries) of $\mathbf{W}$ is a measure of model complexity, and it can be used as a regularizer for maximum likelihood estimation. The main disadvantage of using such penalty is that it leads to a NP-hard problem, regardless of the computational complexity of the log-likelihood.

Next, we formalize the problem of finding a sparse Ising model by regularized maximum likelihood estimation. We replace the cardinality penalty by the $\ell_1$-norm regularizer as in (Wainwright et al., 2006; Banerjee et al., 2008; Höfling & Tibshirani, 2009).

Given a complete dataset with $T$ i.i.d. samples $\mathbf{x}^{(1)}, \ldots, \mathbf{x}^{(T)}$, and a sparseness parameter $\rho > 0$ the $\ell_1$-regularized maximum likelihood estimation for the Ising model in eq.(1) becomes:

$$\min_{\mathbf{W}, \mathbf{b}} \mathcal{L}(\mathbf{W}, \mathbf{b}) + \mathcal{R}(\mathbf{W}) \qquad (2)$$

where the negative (average) log-likelihood $\mathcal{L}(\mathbf{W}, \mathbf{b}) = -\frac{1}{T} \sum_t \log p_{\boldsymbol{\theta}}(\mathbf{x}^{(t)}) = \log \mathcal{Z}(\mathbf{W}, \mathbf{b}) - \langle \widehat{\boldsymbol{\Sigma}}, \mathbf{W} \rangle - \widehat{\boldsymbol{\mu}}^{\mathrm{T}} \mathbf{b}$, the empirical second-order moment $\widehat{\boldsymbol{\Sigma}} = \frac{1}{T} \sum_t \mathbf{x}^{(t)} \mathbf{x}^{(t)\mathrm{T}} - \mathbf{I}$, the empirical first-order moment $\widehat{\boldsymbol{\mu}} = \frac{1}{T} \sum_t \mathbf{x}^{(t)}$ and the regularizer $\mathcal{R}(\mathbf{W}) = \rho \|\mathbf{W}\|_1$.

The objective function in eq.(2) is convex, given the convexity of the log-partition function (Koller & Friedman, 2009), linearity of the scalar products and convexity of the non-smooth $\ell_1$-norm regularizer. As discussed before, computing the partition function $\mathcal{Z}$ is NP-hard, and so is computing the objective function in eq.(2).

## 2.2. Bounds

In what follows, we show boundedness of the optimal solution and the gradients of the maximum likelihood problem. Both are important ingredients for showing convergence and are largely used assumptions in optimization. In this paper, we follow the original formulation of the problem given in (Wainwright et al., 2006; Banerjee et al., 2008; Höfling & Tibshirani, 2009), which does not regularize $\mathbf{b}$. We found interesting to show that this problem has bounds for $\|\mathbf{b}^*\|_1$ unlike other stochastic optimization problems, e.g. SVMs (Shalev-Shwartz et al., 2007).

First, we make some observations that will help us derive our bounds. The empirical second-order moment $\widehat{\boldsymbol{\Sigma}}$ and first-order moment $\widehat{\boldsymbol{\mu}}$ in eq.(2) are computed from binary variables in $\{-1, +1\}$, therefore $\|\widehat{\boldsymbol{\Sigma}}\|_\infty \leq 1$ and $\|\widehat{\boldsymbol{\mu}}\|_\infty \leq 1$.

**Assumption 1.** *It is reasonable to assume that the empirical first-order moment of every variable is not equal to $-1$ (or $+1$), since this would be equivalent to observe a constant value $-1$ (or $+1$) for such variables in every sample in the dataset, i.e.* $(\exists n) \, |\widehat{\mu}_n| = 1 \Leftrightarrow (\forall t) \, x_n^{(t)} = -1 \vee (\forall t) \, x_n^{(t)} = 1$. *Therefore, we assume* $\|\widehat{\boldsymbol{\mu}}\|_\infty < 1 \Leftrightarrow (\forall n) \, -1 < \widehat{\mu}_n < +1$.

Given those observations, we state our bounds in the following theorem. For clarity of the convergence rate analysis, we also define the bound $D$ of the optimal solution.

**Theorem 2.** *The optimal solution* $\boldsymbol{\theta}^* = (\mathbf{W}^*, \mathbf{b}^*)$ *of the maximum likelihood problem in eq.(2) is bounded as follows:*

$$
\begin{aligned}
&\text{i.} && \|\mathbf{W}^*\|_1 \leq \tfrac{N \log 2}{\rho} \\
&\text{ii.} && \|\mathbf{b}^*\|_1 \leq \tfrac{N \log 2 (\rho + 1 + \|\widehat{\boldsymbol{\Sigma}}\|_\infty)}{\rho (1 - \|\widehat{\boldsymbol{\mu}}\|_\infty)} \\
&\text{iii.} && \|\boldsymbol{\theta}^*\|_2 \leq D
\end{aligned}
\qquad (3)
$$

*where* $D^2 = \left( \frac{N \log 2}{\rho} \right)^2 \left( 1 + \left( \frac{\rho + 1 + \|\widehat{\boldsymbol{\Sigma}}\|_\infty}{1 - \|\widehat{\boldsymbol{\mu}}\|_\infty} \right)^2 \right)$.

*Proof Sketch.* Claim i and ii follow from the fact that the function evaluated at $(\mathbf{W}^*, \mathbf{b}^*)$ is less than at $(\mathbf{0}, \mathbf{0})$. Additionally, Claim i follows from non-negativity of the negative log-likelihood in eq.(2), while Claim ii follows from non-negativity of the regularizer and from Assumption 1. Claim iii follows from norm inequalities and Claims i and ii. □

(Please, see Appendix C for detailed proofs.)

If we choose to add the regularizer $\rho \|\mathbf{b}\|_1$ in eq.(2), it is easy to conclude that $\|\mathbf{W}^*\|_1 + \|\mathbf{b}^*\|_1 \leq \frac{N \log 2}{\rho}$ as in Claim i of Theorem 2.

The gradient of the objective function of the maximum likelihood problem in eq.(2) is defined as:

$$
\begin{aligned}
&\text{i.} && \partial \log \mathcal{Z} / \partial \mathbf{W} = \mathbb{E}_{\mathcal{P}}[\mathbf{x}\mathbf{x}^{\mathrm{T}}] \\
&\text{ii.} && \partial \log \mathcal{Z} / \partial \mathbf{b} = \mathbb{E}_{\mathcal{P}}[\mathbf{x}] \\
&\text{iii.} && \partial \mathcal{L} / \partial \mathbf{W} = \partial \log \mathcal{Z} / \partial \mathbf{W} - \widehat{\boldsymbol{\Sigma}} \\
&\text{iv.} && \partial \mathcal{L} / \partial \mathbf{b} = \partial \log \mathcal{Z} / \partial \mathbf{b} - \widehat{\boldsymbol{\mu}}
\end{aligned}
\qquad (4)
$$

where $\mathcal{P}$ is the probability distribution with PMF $p_{\boldsymbol{\theta}}(\mathbf{x})$. The expression in eq.(4) uses the fact that $\mathbb{E}_{\mathcal{P}}[\mathbf{x}\mathbf{x}^{\mathrm{T}}] = \sum_{\mathbf{x}} \mathbf{x}\mathbf{x}^{\mathrm{T}} p_{\boldsymbol{\theta}}(\mathbf{x})$ and $\mathbb{E}_{\mathcal{P}}[\mathbf{x}] = \sum_{\mathbf{x}} \mathbf{x} p_{\boldsymbol{\theta}}(\mathbf{x})$.

It is well known that computing the gradients $\partial \log \mathcal{Z}/\partial \mathbf{W}$ and $\partial \log \mathcal{Z}/\partial \mathbf{b}$ is NP-hard. The complexity results in (Chandrasekaran et al., 2008) imply that approximating these gradients with high probability and arbitrary precision is also NP-hard.

Next, we state some properties of the gradient of the exact log-likelihood. For clarity of the convergence rate analysis, we also define the Lipschitz constant $G$.

**Lemma 3.** *The objective function of the maximum likelihood problem in eq.(2) has the following Lipschitz continuity properties:*

  i. $\|\partial \log \mathcal{Z}/\partial \mathbf{W}\|_\infty$ , $\|\partial \log \mathcal{Z}/\partial \mathbf{b}\|_\infty \leq 1$
  ii. $\|\partial \mathcal{L}/\partial \mathbf{W}\|_\infty \leq 1 + \|\widehat{\mathbf{\Sigma}}\|_\infty$
  iii. $\|\partial \mathcal{L}/\partial \mathbf{b}\|_\infty \;\; \leq 1 + \|\widehat{\boldsymbol{\mu}}\|_\infty$ $\qquad$ (5)
  iv. $\|\partial \mathcal{R}/\partial \mathbf{W}\|_\infty \leq \rho$
  v. $\|\partial \mathcal{L}/\partial \boldsymbol{\theta}\|_2$ , $\|\partial \mathcal{R}/\partial \boldsymbol{\theta}\|_2 \leq G$

*where* $G^2 = N^2 \max((1 + \|\widehat{\mathbf{\Sigma}}\|_\infty)^2 + \frac{1}{N}(1 + \|\widehat{\boldsymbol{\mu}}\|_\infty)^2, \rho^2)$.

*Proof Sketch.* Claims i to iii follow from the fact that the terms $\partial \log \mathcal{Z}/\partial \mathbf{W}$ and $\partial \log \mathcal{Z}/\partial \mathbf{b}$ in eq.(4) are the second and first-order moment of binary variables in $\{-1, +1\}$. Claim iv follows from the definition of subgradients. Claim v follows from norm inequalities and Claims ii to iv. $\qquad \square$

## 2.3. Approximating the Gradient of the Log-Partition Function

Suppose one wants to evaluate the expression $\mathbb{E}_{\mathcal{P}}[\mathbf{x}\mathbf{x}^\mathsf{T}]$ in eq.(4) which is the gradient of the log-partition function. Let assume we know the distribution $p_{\boldsymbol{\theta}}(\mathbf{x})$ up to a constant factor, i.e. $p'_{\boldsymbol{\theta}}(\mathbf{x}) = e^{\mathbf{x}^\mathsf{T}\mathbf{W}\mathbf{x} + \mathbf{b}^\mathsf{T}\mathbf{x}}$. Importance sampling draws $S$ samples $\mathbf{x}^{(1)}, \ldots, \mathbf{x}^{(S)}$ from a trial distribution with PMF $q(\mathbf{x})$, calculates the importance weights $\alpha^{(s)} = p'_{\boldsymbol{\theta}}(\mathbf{x}^{(s)})/q(\mathbf{x}^{(s)})$ and produces the estimate $(\sum_s \alpha^{(s)}\mathbf{x}^{(s)}\mathbf{x}^{(s)\mathsf{T}})/\sum_s \alpha^{(s)}$. On the other hand, MCMC generates $S$ samples $\mathbf{x}^{(1)}, \ldots, \mathbf{x}^{(S)}$ from the distribution $p_{\boldsymbol{\theta}}(\mathbf{x})$ based on constructing a Markov chain whose stationary distribution is $p_{\boldsymbol{\theta}}(\mathbf{x})$. Thus, the estimate becomes $\frac{1}{S}\sum_s \mathbf{x}^{(s)}\mathbf{x}^{(s)\mathsf{T}}$.

In what follows, we characterize a family of samplers that includes importance sampling and MCMC as shown in (Peskun, 1973; Liu, 2001).

**Definition 4.** *A $(B, V, S, D)$-sampler takes $S$ random samples from a distribution $\mathcal{Q}$ and produces biased estimates of the gradient of the log-partition function $\partial \log \mathcal{Z}/\partial \boldsymbol{\theta} + \boldsymbol{\xi}$, with error $\boldsymbol{\xi}$ that has bias and variance:*

  i. $\mathbb{E}_{\mathcal{Q}}[\|\boldsymbol{\xi}\|_2] \;\; \leq \frac{B}{S} + \mathcal{O}(\frac{1}{S^2})$
  ii. $\mathbb{V}\mathrm{ar}_{\mathcal{Q}}[\|\boldsymbol{\xi}\|_2] \leq \frac{V}{S} + \mathcal{O}(\frac{1}{S^2})$ $\qquad$ (6)

*for $B \geq 0, V \geq 0$ and $(\forall \boldsymbol{\theta}) \|\boldsymbol{\theta}\|_2 \leq D$.*

Note that a $(B, V, S, D)$-sampler is asymptotically unbiased with vanishing variance, i.e. $S \to +\infty \Rightarrow \frac{B}{S} \to 0 \land \frac{V}{S} \to 0$. Unfortunately, analytical approximations of the constants $B$ and $V$ are difficult to obtain even for specific classes, e.g. Ising models. The theoretical analysis implies that such constants $B$ and $V$ exist (Peskun, 1973; Liu, 2001) for importance sampling and MCMC. We argue that this apparent disadvantage does not diminish the relevance of our analysis, since we can reasonably expect that more refined samplers lead to lower $B$ and $V$.

Note that Definition 4 does not contradict the complexity results in (Chandrasekaran et al., 2008) that show that it is likely impossible to approximate $\mathcal{Z}$ (and therefore its gradient) with probability greater than $1 - \delta$ and arbitrary precision $\varepsilon$ in time polynomial in $\log \frac{1}{\delta}$ and $\frac{1}{\varepsilon}$. Definition 4 assumes biasedness and a polynomial decay instead of an exponential decay (which is a more stringent condition) and cannot be used to derive two-sided high probability bounds that are both $\mathcal{O}(\log \frac{1}{\delta})$ and $\mathcal{O}(\frac{1}{\delta})$. Therefore, Definition 4 cannot be used to obtain polynomial-time algorithms as the ones considered in (Chandrasekaran et al., 2008).

**Assumption 5.** *It is reasonable to assume that the estimates of the gradient of the log-partition function are inside $[-1; +1]$ since they are approximations of the second and first-order moment of binary variables in $\{-1, +1\}$. Furthermore, it is straightforward to enforce Lipschitz continuity (condition i of Lemma 3) for any sampler (e.g. importance sampling, MCMC or any conceivable method) by limiting its output to be inside $[-1; +1]$. More formally, we have:*

  i. $\|\partial \log \mathcal{Z}/\partial \boldsymbol{\theta} + \boldsymbol{\xi}\|_\infty \leq 1$
  ii. $\|\partial \mathcal{L}/\partial \boldsymbol{\theta} + \boldsymbol{\xi}\|_2 \;\;\;\; \leq G$ $\qquad$ (7)

## 3. Biased Stochastic Optimization

In this section, we analyze the convergence rates of *forward-backward splitting*. Our results apply to any problem that fulfills the following largely used assumptions in optimization:

- the objective function is composed by a smooth function $\mathcal{L}(\boldsymbol{\theta})$ and non-smooth regularizer $\mathcal{R}(\boldsymbol{\theta})$
- the optimal solution is bounded, i.e. $\|\boldsymbol{\theta}^*\|_2 \leq D$
- each visited point is at a bounded distance from the optimal solution, i.e. $(\forall k) \|\boldsymbol{\theta}^{(k)} - \boldsymbol{\theta}^*\|_2 \leq D$
- both $\mathcal{L}$ and $\mathcal{R}$ are Lipschitz continuous, i.e. $\|\partial \mathcal{L}/\partial \boldsymbol{\theta}\|_2$ , $\|\partial \mathcal{R}/\partial \boldsymbol{\theta}\|_2 \leq G$
- the non-smooth regularizer vanishes at zero, i.e. $\mathcal{R}(\mathbf{0}) = 0$

We additionally require that the errors do not change the Lipschitz continuity properties, i.e. $\|\partial \mathcal{L}/\partial \boldsymbol{\theta} + \boldsymbol{\xi}\|_2 \le G$ (as discussed in Assumption 5).

### 3.1. Algorithm

We analyze *forward-backward splitting* (Duchi & Singer, 2009) for deterministic as well as biased stochastic errors, for non-increasing step sizes of the form $\eta_k \in \mathcal{O}(\frac{1}{k^r})$ for $0 < r < 1$. This method is equivalent to basic *proximal gradient* (Schmidt et al., 2011) for $r = 0$ (constant step size). We point out that FBS has $\mathcal{O}(\frac{1}{\sqrt{K}})$ convergence for $r = \frac{1}{2}$, while basic PG has $\mathcal{O}(\frac{1}{K})$ convergence, and accelerated PG has $\mathcal{O}(\frac{1}{K^2})$ convergence. Thus, PG methods have faster convergence but they are more sensitive to errors.

FBS performs gradient descent steps for the smooth part of the objective function, and (closed form) projection steps for the non-smooth part. Here we assume that at each iteration $k$, we approximate the gradient with some (deterministic or biased stochastic) error $\boldsymbol{\xi}^{(k)}$. For our objective function in eq.(2), one iteration of the algorithm is equivalent to:

i. $\boldsymbol{\theta}^{(k+\frac{1}{2})} = \boldsymbol{\theta}^{(k)} - \eta_k(\mathbf{g}_{\mathcal{L}}^{(k)} + \boldsymbol{\xi}^{(k)})$

ii. $\boldsymbol{\theta}^{(k+1)} = \arg\min_{\boldsymbol{\theta}} \left(\frac{1}{2}\|\boldsymbol{\theta} - \boldsymbol{\theta}^{(k+\frac{1}{2})}\|_2^2 + \eta_{k+1}\mathcal{R}(\boldsymbol{\theta})\right)$ (8)

where $\mathbf{g}_{\mathcal{L}}^{(k)} = \frac{\partial \mathcal{L}}{\partial \boldsymbol{\theta}}(\boldsymbol{\theta}^{(k)})$, and $\boldsymbol{\xi}^{(k)}$ is the error in the gradient approximation. Step ii is a projection step for the non-smooth regularizer $\mathcal{R}(\boldsymbol{\theta})$. For the regularizer in our motivating problem $\mathcal{R}(\mathbf{W}) = \rho\|\mathbf{W}\|_1$, Step ii decomposes into $N^2$ independent *lasso* problems.

### 3.2. Convergence Rates for Deterministic Errors

In what follows, we analyze three different flavors of forward-backward splitting: *robust* which outputs the weighted average of all visited points by using the step sizes as in *robust stochastic approximation* (Nemirovski et al., 2009), *basic* which outputs the average of all visited points as in (Duchi & Singer, 2009), or *random* which outputs a point chosen uniformly at random from the visited points. Here we assume that at each iteration $k$, we approximate the gradient with some deterministic error $\boldsymbol{\xi}^{(k)}$. Our results in this subsection will allow us to draw some conclusions regarding not only FBS but also proximal gradient.

In order to make our bounds more general for different choices of step size $\eta_k \in \mathcal{O}(\frac{1}{k^r})$ for some $0 < r < 1$, we use *generalized harmonic numbers* $H_{r,K} = \sum_{k=1}^{K} \frac{1}{k^r}$ and therefore $H_{0,K} = K$, $H_{r,K} \approx \frac{K^{1-r}}{1-r}$ for $0 < r < 1$, $H_{1,K} \approx \log K$ and $H_{r,K} \approx \frac{1-K^{1-r}}{r-1}$ for $r > 1$.

Additionally, we define a weighted error term that will be used for our analysis of deterministic as well as biased stochastic errors. Given a sequence of errors $\boldsymbol{\xi}^{(1)}, \ldots, \boldsymbol{\xi}^{(K)}$ and a set of arbitrary weights $\gamma_k$ such that $\sum_k \gamma_k = 1$, the error term is defined as:

$$A_{\boldsymbol{\gamma},\boldsymbol{\xi}} \equiv \sum_k \gamma_k \|\boldsymbol{\xi}^{(k)}\|_2 \qquad (9)$$

First, we show the convergence rate of robust FBS.

**Theorem 6.** *For a sequence of deterministic errors* $\boldsymbol{\xi}^{(1)}, \ldots, \boldsymbol{\xi}^{(K)}$, *step size* $\eta_k = \frac{\beta}{Gk^r}$ *for* $0 < r < 1$, *initial point* $\boldsymbol{\theta}^{(1)} = \mathbf{0}$, *the objective function evaluated at the weighted average of all visited points converges to the optimal solution with rate:*

$$\mathcal{L}(\overline{\boldsymbol{\theta}}) + \mathcal{R}(\overline{\boldsymbol{\theta}}) - \mathcal{L}(\boldsymbol{\theta}^*) - \mathcal{R}(\boldsymbol{\theta}^*) \le \pi_\eta(K)$$
$$\le \frac{D^2 G}{2\beta H_{r,K}} + 2DA_{\boldsymbol{\gamma},\boldsymbol{\xi}} + \frac{4\beta G H_{2r,K}}{H_{r,K}} \qquad (10)$$

*where* $\overline{\boldsymbol{\theta}} = \frac{\sum_k \eta_k \boldsymbol{\theta}^{(k)}}{\sum_k \eta_k}$, *the weighted average regret* $\pi_\eta(K) = \frac{\sum_k \eta_k (\mathcal{L}(\boldsymbol{\theta}^{(k)}) + \mathcal{R}(\boldsymbol{\theta}^{(k)}))}{\sum_k \eta_k} - \mathcal{L}(\boldsymbol{\theta}^*) - \mathcal{R}(\boldsymbol{\theta}^*)$, *the error term* $A_{\boldsymbol{\gamma},\boldsymbol{\xi}}$ *is defined as in eq.(9), and the error weights* $\gamma_k = \frac{1/k^r}{H_{r,K}}$ *such that* $\sum_k \gamma_k = 1$.

*Proof Sketch.* By Jensen's inequality $\mathcal{L}(\overline{\boldsymbol{\theta}}) + \mathcal{R}(\overline{\boldsymbol{\theta}}) \le \sum_k \eta_k(\mathcal{L}(\boldsymbol{\theta}^{(k)}) + \mathcal{R}(\boldsymbol{\theta}^{(k)}))/\sum_k \eta_k$. Then we apply a technical lemma for bounding consecutive steps (Please, see Appendix B). □

Second, we show the convergence rate of basic FBS.

**Theorem 7.** *For a sequence of deterministic errors* $\boldsymbol{\xi}^{(1)}, \ldots, \boldsymbol{\xi}^{(K)}$, *step size* $\eta_k = \frac{\beta}{Gk^r}$ *for* $0 < r < 1$, *initial point* $\boldsymbol{\theta}^{(1)} = \mathbf{0}$, *the objective function evaluated at the average of all visited points converges to the optimal solution with rate:*

$$\mathcal{L}(\overline{\boldsymbol{\theta}}) + \mathcal{R}(\overline{\boldsymbol{\theta}}) - \mathcal{L}(\boldsymbol{\theta}^*) - \mathcal{R}(\boldsymbol{\theta}^*) \le \pi(K)$$
$$\le \frac{D^2 G(K+1)^r}{2\beta K} + 2^{1+r} D A_{\boldsymbol{\gamma},\boldsymbol{\xi}} + \frac{2^{2+r}\beta G H_{r,K}}{K} \qquad (11)$$

*where* $\overline{\boldsymbol{\theta}} = \frac{\sum_k \boldsymbol{\theta}^{(k)}}{K}$, *the average regret* $\pi(K) = \frac{\sum_k (\mathcal{L}(\boldsymbol{\theta}^{(k)}) + \mathcal{R}(\boldsymbol{\theta}^{(k)}))}{K} - \mathcal{L}(\boldsymbol{\theta}^*) - \mathcal{R}(\boldsymbol{\theta}^*)$, *the error term* $A_{\boldsymbol{\gamma},\boldsymbol{\xi}}$ *is defined as in eq.(9), and the error weights* $\gamma_k = \frac{1}{K}$ *such that* $\sum_k \gamma_k = 1$.

*Proof Sketch.* By Jensen's inequality $\mathcal{L}(\overline{\boldsymbol{\theta}}) + \mathcal{R}(\overline{\boldsymbol{\theta}}) \le \sum_k (\mathcal{L}(\boldsymbol{\theta}^{(k)}) + \mathcal{R}(\boldsymbol{\theta}^{(k)}))/K$. Then we apply a technical lemma for bounding consecutive steps (Please, see Appendix B). □

Finally, we show the convergence rate of random FBS.

**Theorem 8.** *For a sequence of deterministic errors* $\boldsymbol{\xi}^{(1)}, \ldots, \boldsymbol{\xi}^{(K)}$, *step size* $\eta_k = \frac{\beta}{Gk^r}$ *for* $0 < r < 1$, *initial point* $\boldsymbol{\theta}^{(1)} = \mathbf{0}$ *and some confidence parameter* $0 < \varepsilon < 1$, *the objective function evaluated at a point* $k$ *chosen uniformly at random from the visited points converges, with probability at least* $1 - \varepsilon$, *to the optimal solution with rate:*

$$\mathcal{L}(\boldsymbol{\theta}^{(k)}) + \mathcal{R}(\boldsymbol{\theta}^{(k)}) - \mathcal{L}(\boldsymbol{\theta}^*) - \mathcal{R}(\boldsymbol{\theta}^*)$$
$$\leq \frac{1}{\varepsilon} \left( \frac{D^2 G(K+1)^r}{2\beta K} + 2^{1+r} D A_{\boldsymbol{\gamma}, \boldsymbol{\xi}} + \frac{2^{2+r} \beta G H_{r,K}}{K} \right) \quad (12)$$

*where the error term* $A_{\boldsymbol{\gamma}, \boldsymbol{\xi}}$ *is defined as in eq.(9), and the error weights* $\gamma_k = \frac{1}{K}$ *such that* $\sum_k \gamma_k = 1$.

*Proof Sketch.* Since the distribution is uniform on $k$, the expected value of the objective function is equal to the average of the objective function evaluated at all visited points, i.e. the average regret $\pi(K)$. The final result follows from Markov's inequality and the upper bound of $\pi(K)$ given in Theorem 7. □

The convergence rates in Theorems 6, 7 and 8 lead to an error term $A_{\boldsymbol{\gamma}, \boldsymbol{\xi}}$ that is linear, while the error term is quadratic in the analysis of proximal gradient (Schmidt et al., 2011). In basic PG, the error term can be written as:

$$\frac{1}{K} \left( \sum_k \|\boldsymbol{\xi}^{(k)}\|_2 \right)^2 = K (A_{\boldsymbol{\gamma}, \boldsymbol{\xi}})^2 \quad (13)$$

where the error weights $\gamma_k = \frac{1}{K}$ such that $\sum_k \gamma_k = 1$. In accelerated PG, the error term can be written as:

$$\frac{4}{(K+1)^2} \left( \sum_k k \|\boldsymbol{\xi}^{(k)}\|_2 \right)^2 = K^2 (A_{\boldsymbol{\gamma}, \boldsymbol{\xi}})^2 \quad (14)$$

where the error weights $\gamma_k = k / \binom{K}{2}$ so that $\sum_k \gamma_k = 1$.

Note that both PG methods contain terms $K$ and $K^2$, which are not in our analysis. As noted in (Schmidt et al., 2011), errors have a greater effect on the accelerated method than on the basic method. This observation suggests that, unlike in the error-free case, accelerated PG is not necessarily better than the basic method due to a higher sensitivity to errors (Devolder et al., 2011).

Intuitively speaking, basic PG is similar to basic FBS in the sense that errors from all iterations have the same effect on the convergence rate, i.e. $\gamma_k$ is constant. In robust FBS, errors in the last iterations have a lower effect on the convergence rate than errors in the beginning, i.e. $\gamma_k$ is decreasing. In accelerated PG, errors in the last iterations have a bigger effect on the convergence rate than errors in the beginning, i.e. $\gamma_k$ is increasing.

The analysis of Schmidt et al. (2011) for deterministic errors implies that in order to have convergence,

Table 1. Order of errors $\|\boldsymbol{\xi}^{(k)}\|_2$ required to obtain convergence of the error term for the *deterministic* case: basic (PB) and accelerated (PA) proximal gradient, basic (FB) and robust (FR) forward-backward splitting.

| Method | Convergence | | | |
|---|---|---|---|---|
| | for $K \to +\infty$ | $\mathcal{O}(\frac{1}{\sqrt{K}})$ | $\mathcal{O}(\frac{1}{K})$ | $\mathcal{O}(\frac{1}{K^2})$ |
| PB | $\mathcal{O}(\frac{1}{k^{1/2+\epsilon}})$ | $\mathcal{O}(\frac{1}{k^{3/4+\epsilon}})$ | $\mathcal{O}(\frac{1}{k^{1+\epsilon}})$ | - |
| PA | $\mathcal{O}(\frac{1}{k^{1+\epsilon}})$ | $\mathcal{O}(\frac{1}{k^{5/4+\epsilon}})$ | $\mathcal{O}(\frac{1}{k^{3/2+\epsilon}})$ | $\mathcal{O}(\frac{1}{k^{2+\epsilon}})$ |
| FB $(r=\frac{1}{2})$ | $\mathcal{O}(\frac{1}{\log k})$ | $\mathcal{O}(\frac{1}{k^{1/2+\epsilon}})$ | $\mathcal{O}(\frac{1}{k^{1+\epsilon}})$ | - |
| FR $(r=\frac{1}{2})$ | $\mathcal{O}(\frac{1}{\log k})$ | $\mathcal{O}(\frac{1}{k^{1/2+\epsilon}})$ | - | - |

the errors must decrease at a rate $\|\boldsymbol{\xi}^{(k)}\|_2 \in \mathcal{O}(\frac{1}{k^{1/2+\epsilon}})$ for some $\epsilon > 0$ in the case of basic PG, and $\mathcal{O}(\frac{1}{k^{1+\epsilon}})$ for accelerated PG. In contrast, our analysis of FBS show that we only need logarithmically decreasing errors $\mathcal{O}(\frac{1}{\log k})$ in order to have convergence. Regarding $\mathcal{O}(\frac{1}{\sqrt{K}})$ convergence of the error term $A_{\boldsymbol{\gamma}, \boldsymbol{\xi}}$, basic and robust FBS requires errors $\mathcal{O}(\frac{1}{k^{1/2+\epsilon}})$ (the minimum required for convergence in basic PG). Table 1 summarizes the requirements for different convergence rates of the error term $A_{\boldsymbol{\gamma}, \boldsymbol{\xi}}$ of FBS as well as the error terms of basic PG in eq.(13) and accelerated PG in eq.(14).

For an informal (and incomplete) analysis of the results in (Schmidt et al., 2011) for biased stochastic optimization, consider each error bounded by its bias and variance $\|\boldsymbol{\xi}^{(k)}\|_2 \leq B/S_k + c\sqrt{V/S_k}$ for some $c > 0$ and an increasing number of random samples $S_k$ that allows to obtain decreasing errors. Without noting the possible kind of "uniform convergence" of the bound for all $K$ iterations (making $c$ a function of $K$), the number of random samples must increase (at least) at a rate that is quadratic of the rate of the errors. For instance, in order to have $\mathcal{O}(\frac{1}{K})$ convergence, basic PG requires errors to be $\mathcal{O}(\frac{1}{k^{1+\epsilon}})$ and therefore it would require (at least) an increasing number of random samples $S_k \in \mathcal{O}(k^{2+\epsilon})$ for some $\epsilon > 0$. Accelerated PG would require (at least) $S_k \in \mathcal{O}(k^{4+\epsilon})$ in order to obtain $\mathcal{O}(\frac{1}{K^2})$ convergence. If we include the fact that $c$ is a function of $K$, then the required number of random samples would be "worse than quadratic" of the required rate of the errors. Fortunately, a formal analysis in the next subsection shows that this is not the case for all methods except accelerated PG.

### 3.3. Bounding the Error Term for Biased Stochastic Optimization

In what follows, we focus in the analysis of stochastic errors in order to see if better convergence rates can be obtained than the ones informally outlined in the previous subsection. A formal analysis of the er-ror terms show that *forward-backward splitting* for biased stochastic errors requires only a logarithmically increasing number of random samples in order to converge, i.e. $S_k \in \mathcal{O}(\log k)$. More interestingly, we found that the required number of random samples is the same for the deterministic and the biased stochastic setting for FBS and basic PG. On the negative side, we found that accelerated PG is not guaranteed to converge in the biased stochastic setting.

Next, we present our high probability bound for the error term for biased stochastic optimization. One way to bound the error term $A_{\gamma, \xi}$ would be to rely on "uniform convergence" arguments, i.e. to bound the error of each iteration $\|\xi^{(k)}\|_2$ and then use the well-known union bound. We chose to bound the error term itself, by using the fact that errors become independent (but not identically distributed) conditioned to the parameters $\theta^{(1)}, \ldots, \theta^{(K)}$. We also allow for a different number of random samples $S_k$ for each iteration $k$.

**Theorem 9.** *Given $K$ $(B, V, S_k, D)$-samplers each producing estimates with an error $\xi^{(k)}$, and given a set of arbitrary weights $\gamma_k$ such that $\sum_k \gamma_k = 1$. For some confidence parameter $0 < \delta < 1$, with probability at least $1 - \delta$, the error term is bounded as follows:*

$$A_{\gamma, \xi} \le \lambda_1 + \frac{2\sqrt{M}}{3K} \log \frac{1}{\delta} + \sqrt{2\lambda_2 \log \frac{1}{\delta} + \frac{4M}{9K^2} \log^2 \frac{1}{\delta}} \tag{15}$$

*where the bias term $\lambda_1 = \min(2\sqrt{M}, B\sum_k \frac{\gamma_k}{S_k})$ and the variance term $\lambda_2 = \min(4M, V \sum_k \frac{\gamma_k^2}{S_k})$.*

*Proof Sketch.* The bias and variance for each $\|\xi^{(k)}\|_2$ are bounded by $\frac{B}{S_k}$ and $\frac{V}{S_k}$ by Definition 4. By Lemma 3 and Assumption 5 we have $\|\xi^{(k)}\|_2 \le 2\sqrt{M}$ which is the maximum bias, and its square is the maximum variance. By the definition of marginal distribution, we make $\|\xi^{(1)}\|_2, \ldots, \|\xi^{(K)}\|_2$ independent (but not identically distributed) conditioned to the parameters $\theta^{(1)}, \ldots, \theta^{(K)}$. We then invoke Bernstein inequality for properly defined variables such that it applies to the weighted average $A_{\gamma, \xi}$. □

It is interesting to note what happens for a fixed number of random samples $S_k \in \mathcal{O}(1)$. In this case, the bias term $\lambda_1 \in \mathcal{O}(1)$ and therefore FBS will not converge. For robust FBS, the variance term $\lambda_2 \in \mathcal{O}(H_{2r,K}/(H_{r,K})^2)$ which for instance for $r = \frac{1}{2}$ we have $\lambda_2 \in \mathcal{O}(\frac{\log K}{K})$. For basic FBS, the variance term $\lambda_2 \in \mathcal{O}(\frac{1}{K})$. Therefore, for the constant number of random samples, the lack of convergence of FBS is explained only by the bias of the sampler and not its variance.

*Table 2.* Random samples $S_k$ required to obtain convergence of the error term for the *biased stochastic* case: basic (PB) and accelerated (PA) proximal gradient, basic (FB) and robust (FR) forward-backward splitting.

| Method | Convergence | | | |
|---|---|---|---|---|
| | for $K \to +\infty$ | $\mathcal{O}(\frac{1}{\sqrt{K}})$ | $\mathcal{O}(\frac{1}{K})$ | $\mathcal{O}(\frac{1}{K^2})$ |
| PB | $\mathcal{O}(k^{1/2+\epsilon})$ | $\mathcal{O}(k^{3/4+\epsilon})$ | $\mathcal{O}(k^{1+\epsilon})$ | - |
| PA | - | - | - | - |
| FB $(r=\frac{1}{2})$ | $\mathcal{O}(\log k)$ | $\mathcal{O}(k^{1/2+\epsilon})$ | $\mathcal{O}(k^{1+\epsilon})$ | - |
| FR $(r=\frac{1}{2})$ | $\mathcal{O}(\log k)$ | $\mathcal{O}(k^{1/2+\epsilon})$ | - | - |

Table 2 summarizes the requirements for different convergence rates of the error term $A_{\gamma, \xi}$ of FBS as well as the error terms of basic PG in eq.(13) and accelerated PG in eq.(14). Note that convergence for FBS is guaranteed for a logarithmically increasing number of random samples $S_k \in \mathcal{O}(\log k)$. Moreover, in order to obtain convergence rates of $\mathcal{O}(\frac{1}{\sqrt{K}})$ and $\mathcal{O}(\frac{1}{K})$, the required number of random samples is just the inverse of the required rate of the errors for the deterministic case, and not "worse than quadratic" as outlined in our informal analysis of the previous subsection.

One important conclusion from Theorem 9 is that the upper bound of the error term is $\Omega(\frac{1}{K})$ independently of the bias term $\lambda_1$ and the variance term $\lambda_2$. This implies that the error term is $\mathcal{O}(\frac{1}{K})$ for any setting of error weights $\gamma_k$ and number of random samples $S_k$. The main implication is that the error term in accelerated PG in eq.(14) is constant and therefore the accelerated method is not guaranteed to converge.

## 4. Experimental Results

We illustrate our theoretical findings with a small synthetic experiment ($N = 15$ variables) since we want to report the log-likelihood at each iteration. We performed 10 repetitions. For each repetition, we generate edges in the ground truth model $\mathbf{W}_g$ with a 50% density. The weight of each edge is generated uniformly at random from $[-1; +1]$. We set $\mathbf{b}_g = \mathbf{0}$. We finally generate a dataset of 50 samples. We used a "Gibbs sampler" by first finding the mean field distribution and then performing 5 Gibbs iterations. We used a step size factor $\beta = 1$ and regularization parameter $\rho = 1/16$. We also include a two-step algorithm, by first learning the structure by $\ell_1$-regularized logistic regression (Wainwright et al., 2006) and then learning the parameters by using FBS with belief propagation for gradient approximation. We summarize our results in Figure 1.

Our experiments suggest that stochastic optimiza-tion converges to the maximum likelihood estimate. We also show the Kullback-Leibler divergence to the ground truth, and more pronounced effects for importance sampling (Please, see Appendix D).

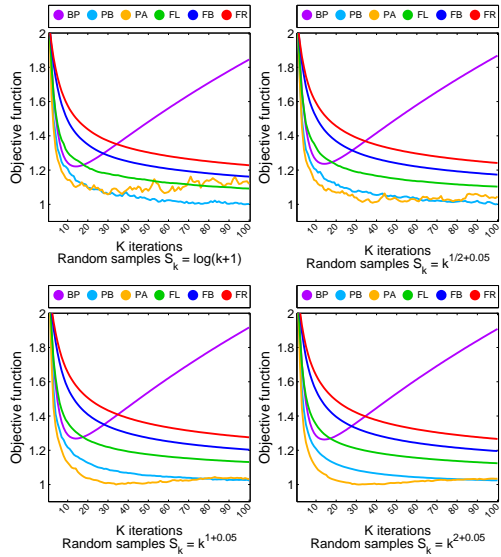

Figure 1. Objective function for different settings of increasing number of random samples. Basic (PB) and accelerated (PA) are noisier and require more samples than last point (FL), basic (FB) and robust (FR) forward-backward splitting in order to converge, but they exhibit faster convergence. Belief propagation (BP) does not converge.

**Concluding Remarks.** Although we focused on Ising models, the ideas developed in the current paper could be applied to Markov random fields with higher order cliques. Our analysis can be easily extended to parameter learning for fixed structures by using a $\ell_2^2$ regularizer instead. Although we show that accelerated proximal gradient is not guaranteed to converge in our specific biased stochastic setting, necessary conditions for its convergence needs to be investigated.

**Acknowledgments.** This work was done while the author was supported in part by NIH Grants 1 R01 DA020949 and 1 R01 EB007530.

## References

Asuncion, A., Liu, Q., Ihler, A., and Smyth, P. Particle filtered MCMC-MLE with connections to contrastive divergence. *ICML*, 2010.

Baes, M. Estimate sequence methods: extensions and approximations. *IFOR internal report, ETH Zurich*, 2009.

Banerjee, O., El Ghaoui, L., and d'Aspremont, A. Model selection through sparse maximum likelihood estimation for multivariate Gaussian or binary data. *JMLR*, 2008.

Barahona, F. On the computational complexity of Ising spin glass models. *Journal of Physics A: Mathematical, Nuclear and General*, 1982.

Besag, J. Statistical analysis of non-lattice data. *The Statistician*, 1975.

Chandrasekaran, V., Srebro, N., and Harsha, P. Complexity of inference in graphical models. *UAI*, 2008.

d'Aspremont, A. Smooth optimization with approximate gradient. *SIAM Journal on Optimization*, 2008.

Desjardins, G., Courville, A., Bengio, Y., Vincent, P., and Delalleau, O. Parallel tempering for training of restricted Boltzmann machines. *AISTATS*, 2010.

Devolder, O. Stochastic first order methods in smooth convex optimization. *CORE Discussion Papers 2012/9*, 2012.

Devolder, O., Glineur, F., and Nesterov, Y. First-order methods of smooth convex optimization with inexact oracle. *CORE Discussion Papers 2011/2*, 2011.

Duchi, J. and Singer, Y. Efficient online and batch learning using forward backward splitting. *JMLR*, 2009.

Duchi, J., Shalev-Shwartz, S., Singer, Y., and Tewari, A. Composite objective mirror descent. *COLT*, 2010.

Duchi, J., Agarwal, A., Johansson, M., and Jordan, M. Ergodic subgradient descent. *Allerton Conference*, 2011.

El Ghaoui, L. and Gueye, A. A convex upper bound on the log-partition function for binary graphical models. *NIPS*, 2008.

Friedlander, M. and Schmidt, M. Hybrid deterministic-stochastic methods for data fitting. *arXiv:1104.2373*, 2011.

Geyer, C. Markov chain Monte Carlo maximum likelihood. *Computing Science and Statistics*, 1991.

Hinton, G. Training products of experts by minimizing contrastive divergence. *Neural Computation*, 2002.

Höfling, H. and Tibshirani, R. Estimation of sparse binary pairwise Markov networks using pseudo-likelihoods. *JMLR*, 2009.

Hu, C., Kowk, J., and Pan, W. Accelerated gradient methods for stochastic optimization and online learning. *NIPS*, 2009.

Jalali, A., Johnson, C., and Ravikumar, P. On learning discrete graphical models using greedy methods. *NIPS*, 2011.

Koller, D. and Friedman, N. *Probabilistic Graphical Models: Principles and Techniques*. The MIT Press, 2009.

Lee, S., Ganapathi, V., and Koller, D. Efficient structure learning of Markov networks using $\ell_1$-regularization. *NIPS*, 2006.

Liu, J. *Monte Carlo Strategies in Scientific Computing*. Springer, 2001.

Marlin, B., Swersky, K., Chen, B., and de Freitas, N. Inductive principles for restricted Boltzmann machine learning. *AISTATS*, 2010.

Murray, I. and Ghahramani, Z. Bayesian learning in undirected graphical models: Approximate MCMC algorithms. *UAI*, 2004.

Nemirovski, A., Juditsky, A., Lan, G., and Shapiro, A. Robust stochastic approximation approach to stochastic programming. *SIAM Journal on Optimization*, 2009.

Parise, S. and Welling, M. Structure learning in Markov random fields. *NIPS*, 2006.

Peskun, P. Optimum Monte Carlo sampling using Markov chains. *Biometrika*, 1973.

Salakhutdinov, R. Learning in Markov random fields using tempered transitions. *NIPS*, 2009.

Salakhutdinov, R. Learning deep Boltzmann machines using adaptive MCMC. *ICML*, 2010.

Schmidt, M., Le Roux, N., and Bach, F. Convergence rates of inexact proximal-gradient methods for convex optimization. *NIPS*, 2011.

Shalev-Shwartz, S., Singer, Y., and Srebro, N. Pegasos: Primal estimated sub-gradient solver for SVM. *ICML*, 2007.

Tieleman, T. Training restricted Boltzmann machines using approximations to the likelihood gradient. *ICML*, 2008.

Wainwright, M., Ravikumar, P., and Lafferty, J. High dimensional graphical model selection using $\ell_1$-regularized logistic regression. *NIPS*, 2006.

Yang, E. and Ravikumar, P. On the use of variational inference for learning discrete graphical models. *ICML*, 2011.

Younes, L. Estimation and annealing for Gibbsian fields. *Annales de l'Institut Henri Poincaré*, 1988.